\documentclass[10pt,twocolumn,letterpaper]{article}

\usepackage{wacv}
\usepackage{times}
\usepackage{epsfig}
\usepackage[accsupp]{axessibility}  

\usepackage{graphicx}
\usepackage{adjustbox}
\usepackage{multirow}
\usepackage{booktabs}
\usepackage{amsmath}
\usepackage{amssymb}
\usepackage{xcolor}

\def\wacvPaperID{630} 

\wacvapplicationstrack 

\wacvfinalcopy 

\ifwacvfinal
\usepackage[breaklinks=true,bookmarks=false]{hyperref}
\else
\usepackage[pagebackref=true,breaklinks=true,colorlinks,bookmarks=false]{hyperref}
\fi

\begin{document}

\title{IDD-3D: Indian Driving Dataset for 3D Unstructured Road Scenes}




\author{Shubham Dokania\textsuperscript{1}, A. H. Abdul Hafez\textsuperscript{2}, Anbumani Subramanian\textsuperscript{1},\\ Manmohan Chandraker\textsuperscript{3}, C.V. Jawahar\textsuperscript{1}\\
\textsuperscript{1}IIIT Hyderabad, \textsuperscript{2}Hasan Kalyoncu University, \textsuperscript{3}UC San Diego\\
{\tt\small shubham.dokania@research.iiit.ac.in, abdul.hafez@hku.edu.tr,}\\
{\tt\small anbumani@iiit.ac.in,  mkchandraker@eng.ucsd.edu, jawahar@iiit.ac.in}
}

\maketitle

\begin{abstract}
   Autonomous driving and assistance systems rely on annotated data from traffic and road scenarios to model and learn the various object relations in complex real-world scenarios. Preparation and training of deploy-able deep learning architectures require the models to be suited to different traffic scenarios and adapt to different situations. Currently, existing datasets, while large-scale, lack such diversities and are geographically biased towards mainly developed cities. An unstructured and complex driving layout found in several developing countries such as India poses a challenge to these models due to the sheer degree of variations in the object types, densities, and locations. To facilitate better research toward accommodating such scenarios, we build a new dataset, {IDD-3D}, which consists of multi-modal data from multiple cameras and LiDAR sensors with 12k annotated driving LiDAR frames across various traffic scenarios. We discuss the need for this dataset through statistical comparisons with existing datasets and highlight benchmarks on standard 3D object detection and tracking tasks in complex layouts. Code and data available \footnote{\url{https://github.com/shubham1810/idd3d_kit.git}}.
\end{abstract}

\section{Introduction}

\begin{figure}[t]
  \centering
  \includegraphics[width=0.99\columnwidth]{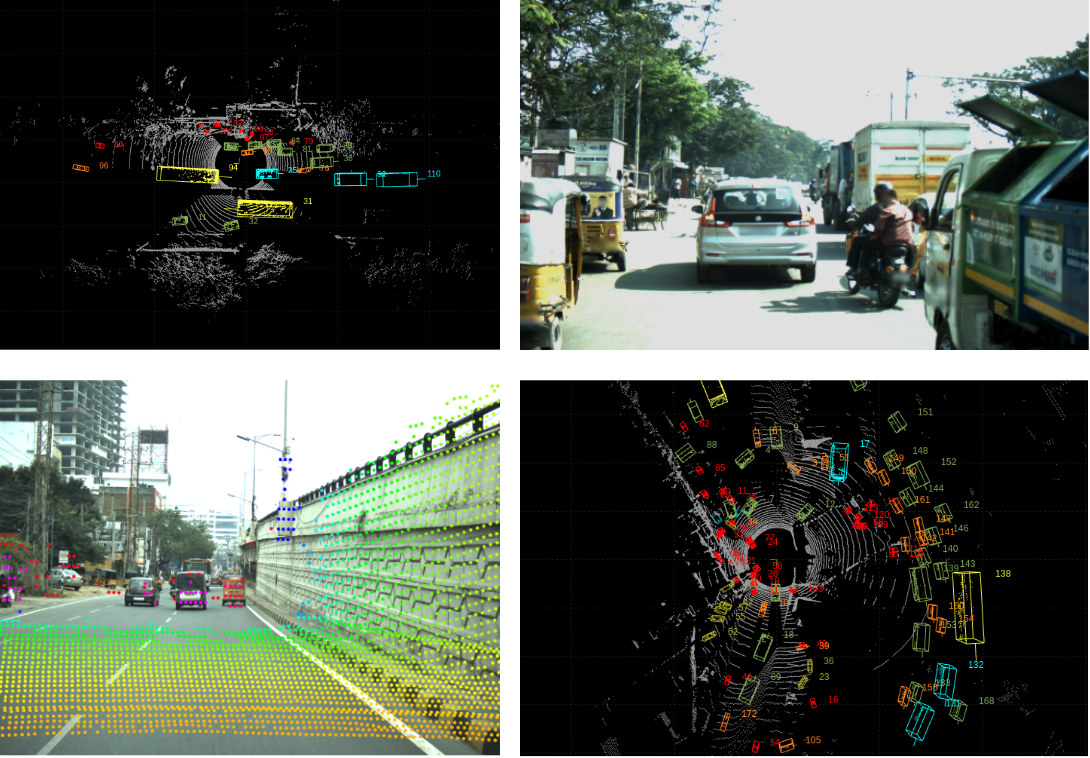}
  \caption{Some examples from the dataset showing different traffic scenarios, LiDAR data with annotations, and a sample of LiDAR point clouds projected on camera data.}
  \label{fig:teaser}
\end{figure}

\begin{figure*}[t]
  \centering
  \includegraphics[width=0.92\linewidth]{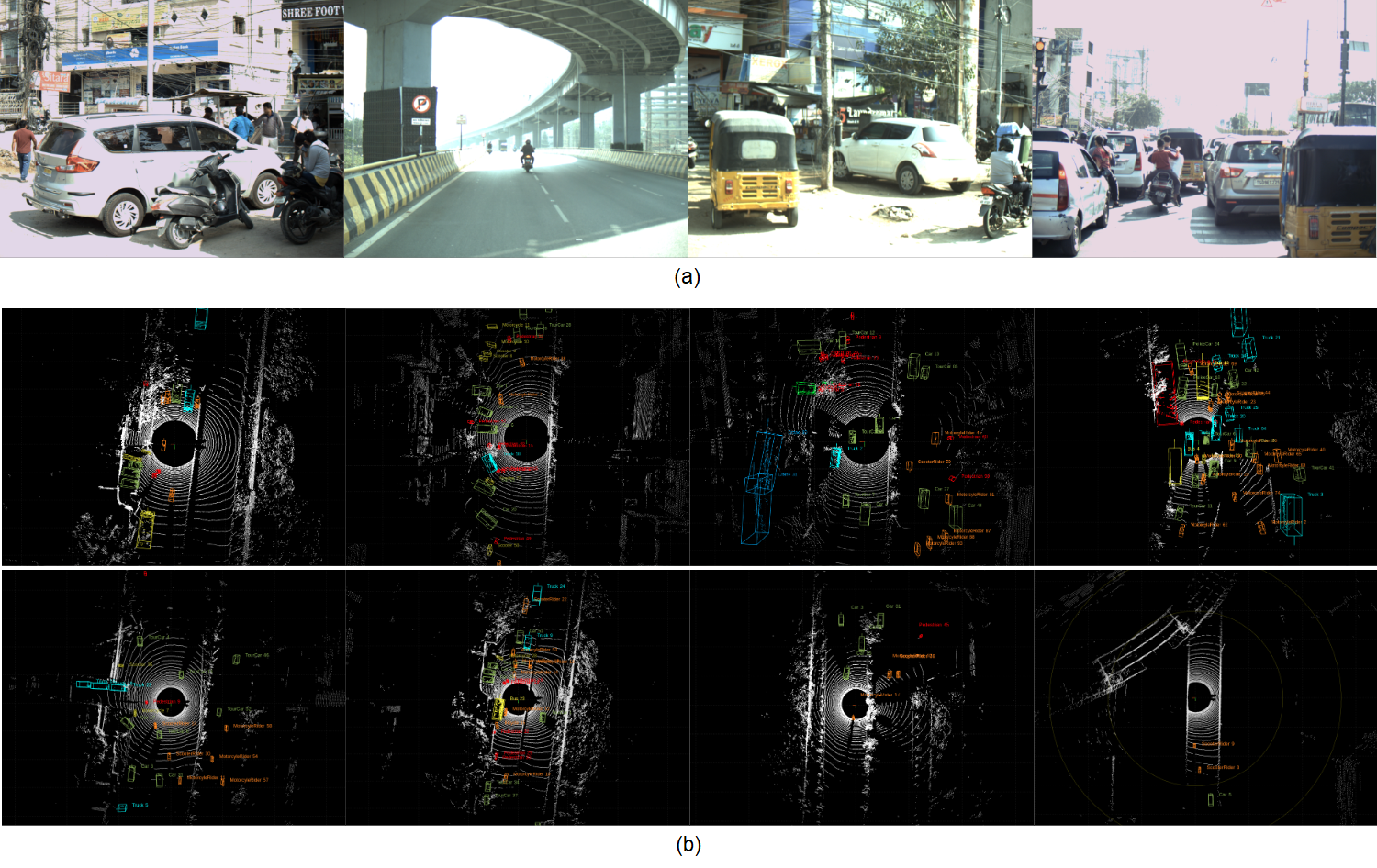}
  \caption{Samples from the dataset highlighting different (a) RGB images and (b) LiDAR Bird-Eye-View (BEV) along with bounding box annotations. The samples visualized above are taken from different sequences of the dataset.}
  \label{fig:samples}
\end{figure*}

Intelligent vehicles and autonomous driving systems have come a long way and keep becoming more sophisticated over time, owing to the rapid progress in the deep learning and computer vision. However, the core component for all these increments is the availability of high-quality annotated data. Recently, many works have focused on data selection and quality improvement \cite{ren2021survey,collins2008towards,yoo2019learning}, building high-quality and large-scale datasets, and approaches built using these resources, which improve the state of autonomous driving \cite{9046805,grigorescu2020survey}.

Existing datasets are usually collected in well-structured environments with proper traffic regulations and relatively-evenly distributed traffic. In such situations, crowd behavior demonstrates low diversity and average densities. In south-east Asian countries, such as India, the traffic densities and inter-object behaviors are much more complex. Such complexities have been studied in the past \cite{8659045,chang2019argoverse,chandra2021meteor}, but extensive data coverage and multi-modal systems are still unavailable for such scenes. It hence may not be entirely applied to cases where the distribution of object categories and types varies greatly.

In this paper, we propose a dataset on complex unstructured driving scenarios with multi-modal data, highlighting the capabilities of 3D sensors such as LiDAR for better scene perception in unstructured and sporadically chaotic traffic conditions. In the proposed dataset, we highlight a significantly different distribution of object types and categories compared to existing datasets collected in European or similar settings \cite{liao2021kitti,gahlert2020cityscapes,sun2020scalability}, due to the different nature of traffic scenes in Indian roads. Furthermore, the categories and annotations available in the proposed dataset vary greatly from existing datasets. Specifically, they cover objects in scenes that usually appear in still-developing cities, for example, Auto-rickshaws, hand carts, concrete mixer machines on roads, and animals on roads. 

\begin{table*}[]
\centering
\begin{adjustbox}{max width=1.8\columnwidth}
\begin{tabular}{lccccccr}
\hline
Dataset                                                     & 3D Scenes & Cameras & Lidar & Images & Classes   & 3D Boxes & Traffic Diversity \\ \hline
KITTI \cite{geiger2012we}                  & 15k       & 2       & yes   & 15k    & 3         & 80k      & Low               \\
nuScenes \cite{nuscenes}                   & 40k       & 6       & yes   & 1.4M   & 23        & 1.4M     & Mid               \\
Apolloscape \cite{8575295}                 & 20k       & 6       & yes   & 0      & 6         & 475k     & Low               \\
KAIST \cite{choi2018kaist}                 & 8.9k      & 2       & yes   & 8.9k   & 3         & 0        & Low               \\
Waymo Open \cite{sun2020scalability}       & 230k      & 5       & yes   & 1M     & 4         & 12M      & Mid               \\
ONCE \cite{mao2021one}                     & 1M (16k)  & 7       & yes   & 7M     & 5         & 417k     & Mid               \\
Cityscapes-3D \cite{gahlert2020cityscapes} & 20k       & -       & no    & 490k   & 8         & -        & Low               \\
A* 3D \cite{pham20203d}                    & 39k       & 1       & yes   & 39k    & 7         & 230k     & Mid               \\ \hline
Ours                                                        & 15.5k*       & 6       & yes   & 93k    & 10 (17**) & 223k*    & High              \\ \hline
\end{tabular}
\end{adjustbox}
\caption{A comparison with existing popular 3D autonomous driving datasets. Our dataset showcases the highest diversity with the highest average number of bounding boxes per frame and a wide distribution. The statistical distribution is further studied in the following sections. (*) Number reported on train-val-test set, experiments/statistics reported on train-val set. (**) The 17 classes are total of the 10 primary and 7 additional classes.}
\label{tab:comparison}
\end{table*}

\begin{figure*}[ht]
  \centering
  \includegraphics[width=0.9\linewidth]{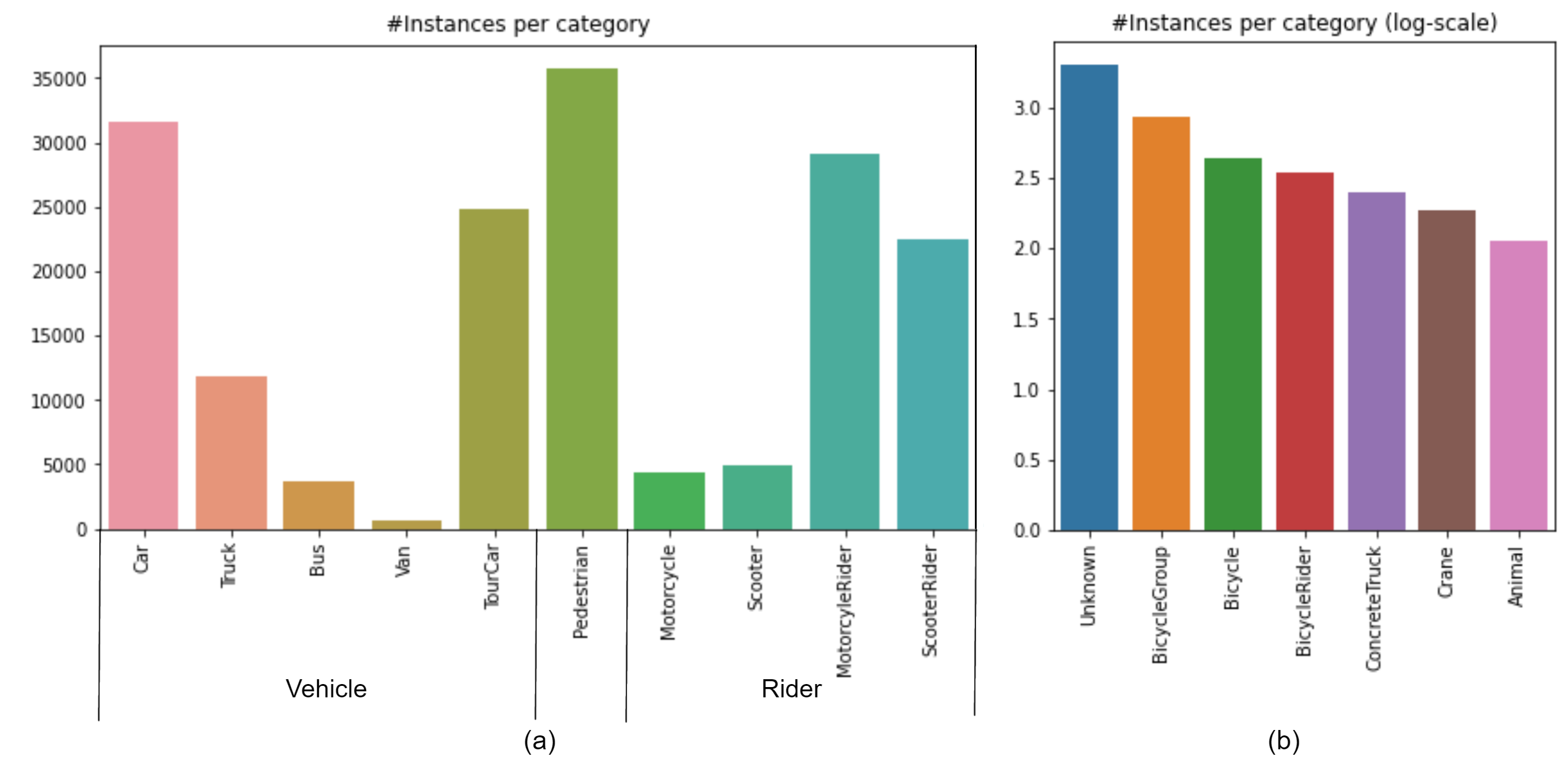}
  \caption{Distribution of class labels in the proposed dataset. (a) The primary 10 classes are shown here along with the 3 super-categories (Vehicle, Pedestrian, and Rider) which are considered to make the proposed dataset more consistent with labels from existing datasets. (b) The additional 7 classes annotated in the dataset are shown in log-scale separately since they are currently not used for training the models. \textit{The Rider class covers both riders and non-riders on two-wheeler motor vehicles.} We do not consider the Miscellaneous classes for evaluation of the dataset currently.}
  \label{fig:stats}
\end{figure*}

We provide data collected in Indian road scenes, from high-quality LiDAR sensors and six cameras that cover the surrounding area of the ego-vehicle to enable sensor-fusion-based applications. We provide annotations for 15.5k frames in the dataset, which spans 10 primary categories (and 7 additional miscellaneous categories), which we use for model training and evaluation. Along with the annotations, we also provide extra unlabelled raw data from the sensors to facilitate further research, especially into self- and unsupervised learning over such traffic scenes. A unique feature of the proposed dataset, which stems from the unstructured environment, is the availability of highly complex trajectories. We show samples from the dataset which emphasize such cases and display experiments on object detection and tracking, which is possible due to availability of instance specific labels for each object bounding box per sequence.

Our main contributions can be summarised as follows: (i) We propose the IDD-3D dataset for driving in unstructured traffic scenarios for Indian roads with 3D information, (ii) high-quality annotations for 3D object bounding boxes with 9DoF data, and instance IDs to enable tracking, (iii) Analysis over highly unstructured and diverse environments to accentuate the usefulness of proposed dataset, and (iv) provide 3D object detection and tracking benchmarks across popular methods in literature.

\section{Related Work}

Data plays a huge role in machine learning systems, and in this context, for autonomous vehicles and scene perception. There have been several efforts over the years in this area to improve the state of datasets available and towards increasing the volumes of high-quality and well annotated datasets.

\paragraph{2D Driving:} One of the early datasets towards visual perception and understanding driving has been the CamVid \cite{brostow2009semantic} and Cityscapes \cite{Cordts2016Cityscapes,Cordts2015Cvprw} dataset, providing annotations for semantic segmentation and enabling research in deeper scene understanding at pixel-level. KITTI \cite{geiger2013vision,geiger2012we} dataset provided 2D object annotations for detection and tracking along with segmentation data. However, fusion of multiple modalities such as 3D LiDAR data enhances the performance for scene understanding benchmarks as these provide a higher level of detail of a scene when combined with available 2D data. This multi-modal sensor-fusion based direction has been the motivation for the proposed dataset to alleviate the discrepancies in existing datasets for scene perception and autonomous driving.


\begin{figure*}[ht]
  \centering
  \includegraphics[width=0.95\linewidth]{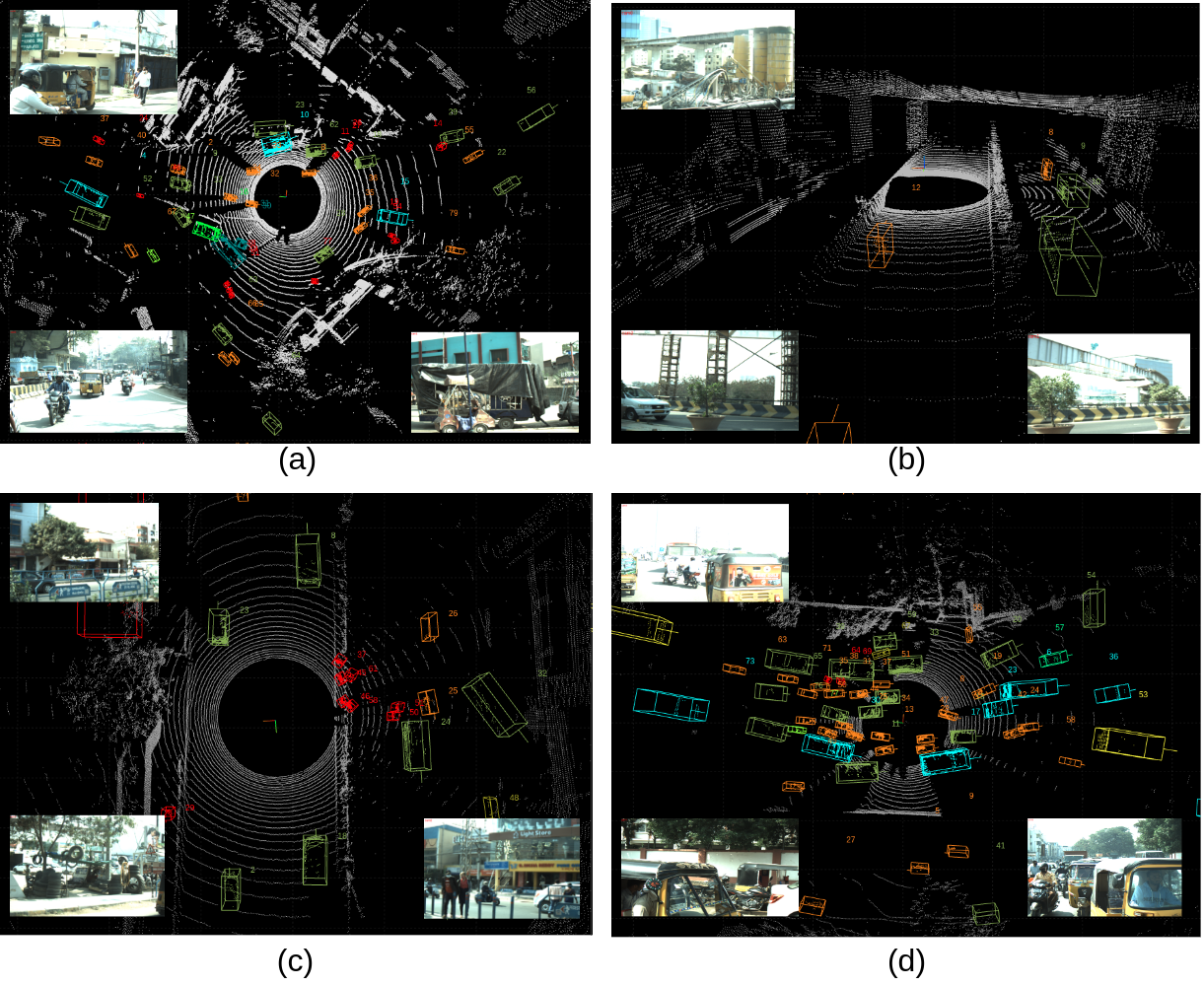}
  \caption{Samples of scenes of interest in our dataset (LiDAR and RGB samples) which especially differentiate our proposed dataset from those available in literature. \textit{(Clockwise from top-left)} (a) Complex traffic scenarios with vehicles orientations in a wide variety of directions, (b) Perspective view of a scene with ego-vehicle on elevated flyover with ground level visible and another highway over the vehicle path with pillars, (c) humans in the middle of traffic (shown in red boxes) and jaywalking near moving vehicles, resulting in a safety critical scenario, (d) An example with very high density traffic scenario. Such case are abundant in the proposed dataset (rather than special cases when compared to other popular datasets) and hence require special attention for such unstructured environments.  \textit{Refer to supplementary material for more examples.}}
  \label{fig:interesting}
\end{figure*}

\begin{figure*}[t]
  \centering
  \includegraphics[width=0.98\linewidth]{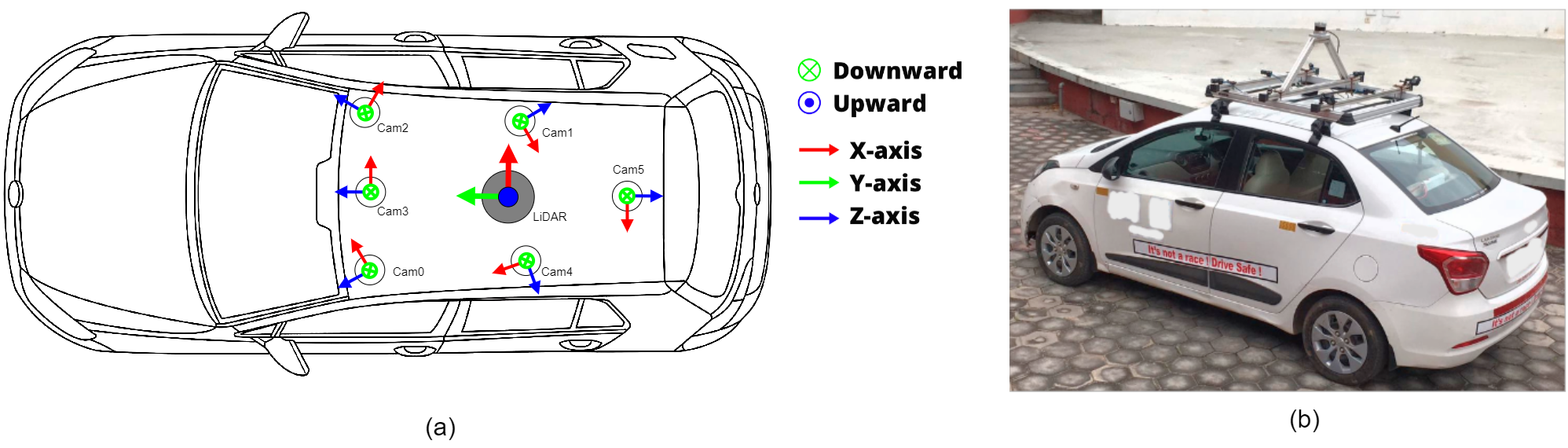}
  \caption{Figure showing (a) sensors on the vehicle (cameras, LiDAR) and their respective orientations, (b) image of the vehicle used along with the sensor rig. \textit{Please note that the real-world car image has been edited to preserve anonymity.}}
  \label{fig:sensors}
\end{figure*}

\paragraph{Driving Datasets:} Recent datasets such as nuScenes \cite{nuscenes}, Argoverse \cite{chang2019argoverse}, Argoverse 2 \cite{wilson2021argoverse} provide HD maps for road scenes. This allows for improved perception and planning capabilities and towards construction of better metrics for object detection such as in \cite{sun2020scalability}. These large scale datasets cover a variety of scenes and traffic densities and have enabled systems with high safety regulations in the area of driver assistance and autonomous driving. However, the drawback for a majority of these datasets arises from the fact that the collection happens in well-developed cities with clear and structured traffic flows. The proposed dataset bridges the gaps of varying environments by introducing more complex environments and extending the diversity of driving datasets.

\paragraph{Complex environments:} There have been multiple efforts to build datasets for difficult environments such as variations in extreme weather \cite{pitropov2021canadian,sakaridis2021acdc}, night-time driving conditions \cite{dai2018dark}, and safety critical scenarios \cite{bao2020uncertainty}. There have been recent works which make use of different sensors such as fisheye lenses to cover a larger area around the ego-vehicle \cite{yogamani2019woodscape,liao2021kitti} and event camera \cite{rebecq2019high} for training models with faster reaction times. However, most of these datasets have been collected in environments with little to no changes in the traffic patterns and consistency in the background objects. Some works in literature \cite{8659045,sharma2022cat,jiang2021rellis,wigness2019rugd} explore such situations where the label distributions can vary significantly, however these are either limited to mostly 2D modalities, or off-road environments. In this work, the proposed dataset enhances the availability of data for enabling research for autonomous driving in unconstrained traffic environments.

\paragraph{Object Detection and Tracking:} Several popular methods have been explored in recent literature which handle the task of 3D object detection for the cases of driving scenarios \cite{zhou2018voxelnet,yang2018pixor,yan2018second,lang2019pointpillars,yin2021center}. In our work, we specifically talk about 3D object detection from point clouds, while we do note the effectiveness of multi-modal approaches as well \cite{chen2017multi,qi2018frustum,sindagi2019mvx}. We have used approaches such as SECOND \cite{yan2018second} which voxelize the input point cloud and apply 3D convolution, which leads to discrete geometric representations of the data. CenterPoint \cite{yin2021center} approach which assigns centers is known to perform well for smaller objects due to the fine level of details for each point feature. We also explore PointPillars \cite{lang2019pointpillars} for an analysis of pillar based approaches where the data is projected to Bird-Eye-View mode and then treated as an image. We highlight the performance of each in the experiments section and draw our inferences specific to the proposed dataset.

Many methods have been proposed towards 3D Multi-Object Tracking (MOT) in literature which have been shown to perform well across a multitude of datasets in different scenarios. There are various ways to model the tracking task such as using the Bird-Eye View \cite{liu2022bevfusion}, approaches based on multi-sensor fusion \cite{kim2021eagermot}, and simple tracking based on distance metrics and methods like Kalman filter \cite{pang2021simpletrack}. In this work, we utilise the method presented in \cite{pang2021simpletrack} using the detections from our trained models on {IDD-3D} and present the evaluations based on popular MOT metrics such as the ones presented in \cite{weng20203d}.

\begin{figure}[ht]
  \centering
  \includegraphics[width=0.99\linewidth]{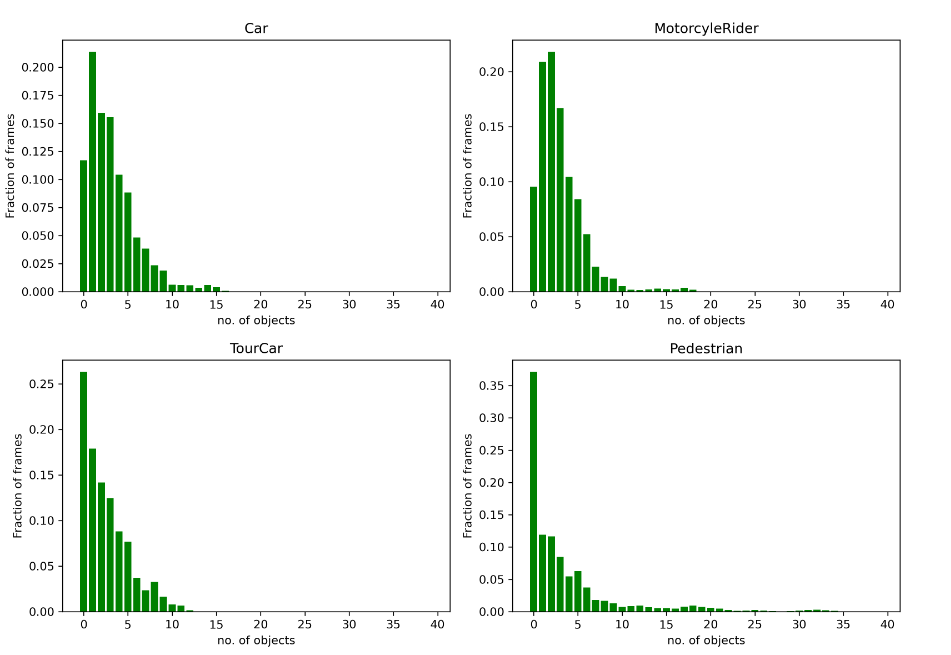}
  \caption{Class-wise distribution of some common prominent classes (Car, MotorcycleRider, Pedestrian, TourCar) with respect to number of frames is visualized to show traffic and crowd density in proposed dataset. \textit{Distributions for all classes is shown in supplementary material.}}
  \label{fig:class_stats}
\end{figure}

\begin{figure}[ht]
  \centering
  \includegraphics[width=0.95\linewidth]{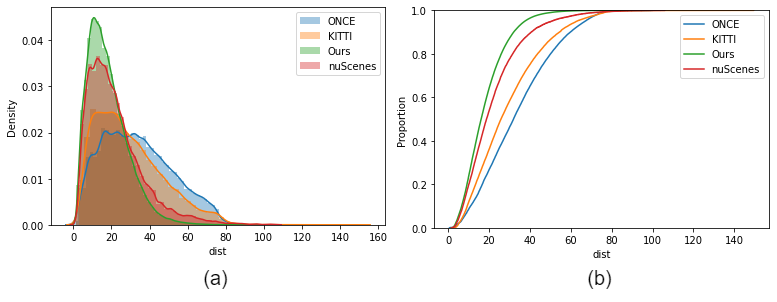}
  \caption{(a) Distribution showing distances of all bounding boxes from the ego-vehicle. The short distance of vehicles and pedestrians provides motivation for the proposed dataset to facilitate modeling of shorter reaction times. (b) Cumulative distribution of the distances further highlight the differences in distance distributions, showing that most of the objects in the proposed dataset are close to the ego-vehicle compared to existing popular datasets.}
  \label{fig:dist_stats}
\end{figure}

\begin{figure}[ht]
  \centering
  \includegraphics[width=0.99\linewidth]{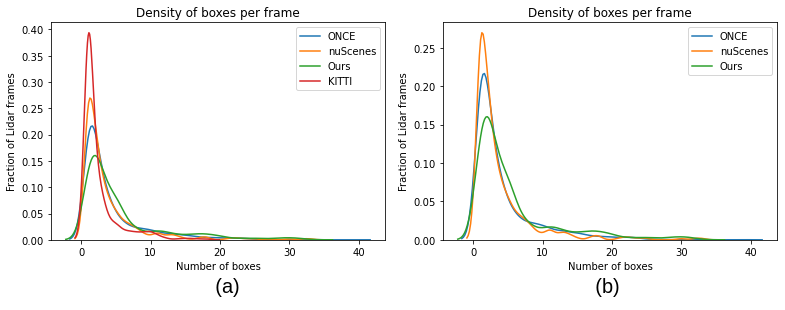}
  \caption{Distributions of number of bounding boxes per LiDAR frame. The number of objects in a scene is usually higher in the frames present in the proposed dataset. We filter the boxes specifically based on the distance of less than 30m based on data shown in fig \ref{fig:class_stats}. (a) Shows statistics with KITTI dataset, and (b) shows the same without KITTI dataset to highlight the sparsity in the KITTI dataset. We note the heavier tail of our distribution indicating a greater density of objects close to the ego-vehicle.}
  \label{fig:box_stats}
\end{figure}

\section{Proposed Dataset}

In the following sections we discuss and highlight the qualities of the proposed dataset, including the design choices and method for data collection, annotations and analysis of the dataset over interesting scenarios.

\subsection{Data Acquisition}

The data collection for the proposed dataset was covered in two driving sessions with over 5 hours of collected data during daytime. Afterwards, we manually sample scenes of interest in sequences of 100 frames at 10fps making 150 sequences, each of 10s. The data collection has been performed in different regions of Hyderabad, India. We now provide details about the configuration and data preparation in the following.  

\paragraph{Sensors (Hardware configuration):}  The proposed dataset encompasses data from multiple sensors which include six RGB cameras and one LiDAR (Ouster OS1) sensor. The details about the sensors and data processing used are mentioned in Table 1 in supplementary material. The position and orientation of the sensors on the acquisition vehicle is shown in Fig. \ref{fig:sensors} along with the real-world image of the vehicle.

\paragraph{Data processing: } For each driving sequence, all calibrations are performed through popular methods such as \cite{iyer2018calibnet,pandey2012automatic}. We preserve the raw data from the sensors in rosbag format \cite{quigley2009ros}. The current release of the dataset consists of 15.5k total frames out of which 12k frames are from train-val set.

\paragraph{Data Privacy:} We ensure that all the faces and license plates in the dataset are blurred by first using automated approaches (such as \cite{deng2020retinaface,laroca2021efficient}) and then performing a manual quality inspection. For the automated approaches, we run the object detection pipeline and then perform a NMS based matching to find any missing boxes in between frames. The missing boxes are interpolated, and finally, we blur the regions in the images for data protection.

\begin{table*}
\centering
\begin{tabular}{@{}lcccccr@{}}
\toprule
SuperCategory            & Categories/Methods & CenterPoint   & \begin{tabular}[c]{@{}c@{}}CenterPoint\\ (nuScenes)\end{tabular} & SECOND         & \begin{tabular}[c]{@{}c@{}}SECOND\\ (KITTI)\end{tabular} & PointPillar \\ \midrule
\multirow{5}{*}{Vehicle} & Car                & 65.28          & 66.97                                                             & \textbf{69.89} & 68.50                                                    & 67.77        \\
                         & Bus                & 59.09          & \textbf{78.47}                                                    & 59.12          & 49.69                                                    & 43.70        \\
                         & Truck              & 68.79          & \textbf{72.18}                                                    & 65.11          & 68.09                                                    & 63.68        \\
                         & Van                & 9.58           & 12.71                                                             & 1.27           & \textbf{15.77}                                           & 0.14         \\
                         & TourCar            & 76.94          & \textbf{77.40}                                                    & 74.81          & 77.02                                                    & 72.80        \\ \midrule
Pedestrian               & Pedestrian         & \textbf{28.60} & 22.49                                                             & 19.54          & 23.74                                                    & 22.72        \\ \midrule
\multirow{4}{*}{Rider}   & Motorcycle         & 23.65          & \textbf{25.28}                                                    & 21.69          & 22.79                                                    & 16.97        \\
                         & Scooter            & \textbf{42.36} & 38.05                                                             & 26.98          & 23.73                                                    & 16.81        \\
                         & MotorcycleRider    & 59.29          & \textbf{61.48}                                                    & 53.39          & 48.90                                                    & 46.52        \\
                         & ScooterRider       & \textbf{66.33} & 64.65                                                             & 52.27          & 50.62                                                    & 41.60        \\ \midrule
                         & mAP                & 49.99          & \textbf{51.97}                                                    & 44.31          & 44.89                                                    & 39.27        \\ \bottomrule
\end{tabular}
\caption{Results on {IDD-3D} with popular methods. We report AP scores across different categories on the validation set. This table shows the results on each training class. The scores are reported with different thresholds for each class (Vehicles @ 0.5, Rider @ 0.4, Pedestrian @ 0.3) and all objects are considered till 30m distance, please see supplementary material for more details and full table.}
\label{tab:exp1}
\end{table*}

\begin{table*}
\centering
\begin{adjustbox}{max width=\linewidth}
\begin{tabular}{lccccccccc}
\hline
\multicolumn{1}{c}{\multirow{2}{*}{Approach}} & \multirow{2}{*}{Pre-Training} & \multicolumn{4}{c}{Vehicle}                                          & \multicolumn{4}{c}{Rider}                                            \\ \cline{3-10} 
\multicolumn{1}{c}{}                          &                               & Overall        & 0-10m          & 10-25m         & \textgreater{}25m & Overall        & 0-10m          & 10-25m         & \textgreater{}25m \\ \hline
CenterPoint                                   & nuScenes                      & \textbf{73.85} & 87.57          & \textbf{70.98} & \textbf{30.48}    & 71.03          & 84.24          & 69.54          & 23.42             \\
CenterPoint                                   & -                             & 71.20          & \textbf{88.84} & 67.62          & 26.32             & 69.51          & 83.66          & 67.49          & 19.76             \\
SECOND                                        & KITTI                         & 72.51          & 88.60          & 68.99          & 28.07             & 71.60          & 83.25          & \textbf{70.98} & 24.32             \\
SECOND                                        & -                             & 73.01          & 88.71          & 67.82          & 29.46             & \textbf{72.05} & \textbf{85.44} & 70.89          & \textbf{26.28}    \\
PointPillar                                   & -                             & 68.61          & 87.64          & 64.59          & 26.30             & 69.66          & 82.56          & 68.60          & 25.64             \\ \hline
\end{tabular}
\end{adjustbox}
\caption{Experimental results on proposed dataset with different popular methods. We report AP scores across different categories on the validation set. This table shows the results on Vehicle and Rider categories from the proposed dataset.}
\label{tab:exp2}
\end{table*}

\begin{table*}
\centering
\begin{adjustbox}{max width=\linewidth}
\begin{tabular}{lccccccccc}
\hline
\multicolumn{1}{c}{\multirow{2}{*}{Approach}} & \multicolumn{1}{l}{\multirow{2}{*}{Pre-Training}} & \multicolumn{4}{c}{Pedestrian}                                       & \multicolumn{4}{c}{mAP}                                              \\ \cline{3-10} 
\multicolumn{1}{c}{}                          & \multicolumn{1}{l}{}                              & Overall        & 0-10m          & 10-25m         & \textgreater{}25m & Overall        & 0-10m          & 10-25m         & \textgreater{}25m \\ \hline
CenterPoint                                   & nuScenes                                          & 22.49          & 33.85          & 19.47          & 4.48              & 55.79          & 68.56          & 53.33          & 19.46             \\
CenterPoint                                   & -                                                 & \textbf{28.60} & \textbf{44.89} & \textbf{24.39} & 3.48              & \textbf{56.43} & \textbf{72.46} & 53.17          & 16.52             \\
SECOND                                        & KITTI                                             & 23.74          & 33.67          & 21.05          & 5.58              & 55.95          & 68.51          & \textbf{53.67} & 19.32             \\
SECOND                                        & -                                                 & 19.54          & 27.18          & 17.61          & \textbf{6.44}     & 54.87          & 67.11          & 52.11          & \textbf{20.73}    \\
PointPillar                                   & -                                                 & 22.72          & 29.34          & 20.45          & 5.45              & 53.66          & 66.52          & 51.21          & 19.13             \\ \hline
\end{tabular}
\end{adjustbox}
\caption{Experimental results (continued) on proposed dataset with different popular methods. We report AP scores across different categories on the validation set. This table shows the results on Pedestrian category and the mAP score from the proposed dataset.}
\label{tab:exp3}
\end{table*}

\subsection{Dataset Analysis}

\paragraph{Labels and Annotations} We provide 3D bounding box annotations for 15.5k (train-val-test) LiDAR frames with ~223k 3D bounding boxes. We have used the annotation tool \cite{li2020sustech} for labeling data across {17} categories, shown in Fig. \ref{fig:stats}. Each object in a sequence contains a unique ID which enables tracking and re-identification. Furthermore, we provide class specific object distribution based on number of frames for some of the prominent categories in Fig. \ref{fig:class_stats}. We note that out of the 17 available classes (primary and additional), we are using 10 primary classes currently for training and validation is performed on 10 classes and 3 super-classes (Vehicle, Pedestrian, and Rider).

\paragraph{Data Statistics}: We first highlight the bounding box distance distribution in {IDD-3D} and the comparison with existing popular datasets \cite{nuscenes,geiger2012we,mao2021one} in figure \ref{fig:dist_stats}. In Fig. \ref{fig:dist_stats} (a) we show that {IDD-3D} consists of most of the annotations close to the ego-vehicle, caused by the low gaps between vehicles causing occlusion for LiDAR rays for longer distances. Nonetheless, it is crucial to highlight this feature of the proposed dataset because split-second decisions are important for safety, especially when other objects are close to the ego-vehicle. We also show better data density compared to KITTI, which is on a comparable scale to {IDD-3D}. Additionally, it can be seen that in the range of 0-25m (where most of the proposed dataset's annotations exist), we show higher densities than both ONCE and nuScenes as shown in fig. \ref{fig:box_stats}(b).

\paragraph{Interesting cases:} While existing datasets provide high diversity in type of traffic scenarios, these are usually restricted to controlled and well-structured environments with only a few anomalies. In {IDD-3D}, we show a large amount of diversity in the situations and also highlight some cases which could be of interest for progress in driving behaviour modeling such as the samples shown in Fig. \ref{fig:interesting}. For example, we see safety critical cases where multiple pedestrians are seen jaywalking while vehicles are on the roads. Existing datasets claim high density traffic when there are 20-30 object bounding boxes in one frame, whereas in our samples we show 50-60 or more objects existing in the same frame, and in close proximity. 
Considering the different variations of scenes in the proposed dataset, the applications for surveillance, road-safety, traffic quality, and crowd-behaviour are immense and show potential to be disparate from the data patterns from other datasets.

\section{Experiments and Benchmarks}

We present an extensive analysis of {IDD-3D} with existing methods to highlight the diversity and usefulness data. We first discuss the experimental setup and then based on the evaluations, report the understanding about the dataset properties and behaviour of different approaches.

\paragraph{Proposed Dataset:} We use 10 primary categories which are highlighted in Fig. \ref{fig:stats}, however, since most datasets in literature ordinarily provide a few categories as common labels (For example, Car, truck, Van as Vehicle), we combine our class labels into three categories, namely Vehicle, Pedestrian, and Rider as super-categories. The network architectures are trained on 10 categories (Car, Bus, Truck, Scooter, Van, Motorcycle, Pedestrian, MotorcycleRider, ScooterRider, TourCar). We transform the annotations to a simpler format for the 3D object detection task a 7-dimensional vector as $(x, y, z, w, h, l, \alpha)$, where $(x, y, z)$ represent the object location, $(w, h, l)$ represent the dimensions of the bounding box and $\alpha$ represents the yaw angle.

\paragraph{3D Object detection:} We discuss about some of the popular datasets which have been considered for comparison with the proposed dataset and highlight their strengths and weaknesses in the complex setting of the presented driving scenarios. For fair comparison, we train network architectures proposed in \cite{yan2018second,yin2021center,lang2019pointpillars} for 3D object detection and show the results in Tables \ref{tab:exp1}, \ref{tab:exp2} and \ref{tab:exp3}. We report the mAP scores for the 3 combined categories (Vehicle, Pedestrian, and Rider) in Tables \ref{tab:exp2} and \ref{tab:exp3}, and further report mAP scores in four sub-levels, i.e. overall AP score for each training class in Table \ref{tab:exp1}. The scores reported in Table \ref{tab:exp1} are for a distance up to 30m in the dataset, and the distances in the super-classes are divided as upto 30m (denoted as Overall), 0-10m, 10-25m, and 25+m. The small distance buckets are considered due to the data distribution (as shown in Fig. \ref{fig:dist_stats}) in the proposed dataset. 

\paragraph{3D Object Tracking:} A notable property of the proposed dataset is the existence of the instance IDs for each 3D bounding box. In this work, we also show results on 3D object tracking and report important metrics such as AMOTA, AMOTP \cite{weng20203d} in Table \ref{tab:exp4}. We use SimpleTrack \cite{pang2021simpletrack} for the task of object tracking and report the results based on the detections from Centerpoint \cite{yin2021center} due to the highest mAP score on the detection task. The MOT scores are reported for all 10 primary classes and the overall categories.

\paragraph{Datasets}: We use KITTI \cite{geiger2012we,geiger2013vision} dataset and nuScenes \cite{nuscenes} for pre-training of 3D object detection methods to further fine-tune on our proposed dataset. We note that cross-dataset training may not be fruitful in this scenario given the significantly different distribution of the categories and input data in the given datasets. The existing datasets usually utilise information such as LiDAR intensity, elongation, and timestamp information as input to the model, which is different from the proposed dataset. However, considering the wide research available based on these datasets, it is imperative that we highlight how using the existing models trained on these datasets as pre-training backbones usually enhances the performances. For this purpose, we consider using the models \cite{yan2018second,yin2021center} for pre-training by using the weights for the common layers and fine-tune for better performance.

\begin{table*}
\centering
\begin{adjustbox}{max width=\linewidth}
\begin{tabular}{@{}lccccccccr@{}}
\toprule
\textbf{Category} & \textbf{AMOTA} & \textbf{AMOTP} & \textbf{Recall} & \textbf{MOTAR} & \textbf{MOTP} & \textbf{MOTA} & \textbf{lgd} & \textbf{tid} & \textbf{faf} \\ \midrule
Bus               & 0.831          & 0.679          & 0.812           & 0.907          & 0.589         & 0.736         & 3.045        & 2.659        & 13.805       \\
Car               & 0.641          & 0.726          & 0.667           & 0.787          & 0.518         & 0.521         & 3.422        & 2.035        & 44.806       \\
Motorcycle        & 0.202          & 0.826          & 0.242           & 0.941          & 0.356         & 0.228         & 2.000        & 2.000        & 2.321        \\
MotorcyleRider    & 0.507          & 0.735          & 0.496           & 0.801          & 0.320         & 0.390         & 5.027        & 2.585        & 36.410       \\
Pedestrian        & 0.254          & 0.912          & 0.319           & 0.737          & 0.363         & 0.225         & 9.918        & 6.731        & 34.557       \\
Scooter           & 0.250          & 0.494          & 0.323           & 1.000          & 0.092         & 0.323         & 0.000        & 0.000        & 0.000        \\
ScooterRider      & 0.540          & 0.536          & 0.581           & 0.742          & 0.258         & 0.427         & 3.868        & 2.274        & 35.251       \\
TourCar           & 0.796          & 0.433          & 0.848           & 0.821          & 0.351         & 0.692         & 2.877        & 1.034        & 48.866       \\
Truck             & 0.701          & 0.635          & 0.675           & 0.903          & 0.403         & 0.607         & 5.108        & 2.676        & 17.796       \\
Van               & 0.000          & 1.677          & 0.275           & 0.000          & 0.563         & 0.000         & 14.500       & 0.000        & 75.163       \\ \midrule
\textbf{Overall}  & 0.472          & 0.765          & 0.524           & 0.764          & 0.381         & 0.415         & 4.977        & 2.199        & 30.898       \\ \bottomrule
\end{tabular}
\end{adjustbox}
\caption{Experimental results for 3D object tracking for the 10 primary classes present in the proposed dataset. We use SimpleTrack \cite{pang2021simpletrack} for the task of tracking using detections from CenterPoint \cite{yin2021center} in the presented table. For the abalation study, please refer to the supplementary material.}
\label{tab:exp4}
\end{table*}

\paragraph{Result Analysis}: We note that the performance of the architectures for both 10 categories and the 3 super-categories is consistent and aligns with our claims. It is clear that the number of annotated instances plays a major role for better mAP scores, for example, classes such as Car achieve a high mAP compared to classes such as Van or Scooter. Another major factor appears to be the object size, wherein larger and denser objects are easier to model and detect compared to smaller instances. An example of the variations in mAP scores based on sizes is the differences between the Pedestrian and Bus/Truck categories, even though Pedestrian category consists of the maximum bounding box instances. From Table \ref{tab:exp1}, we see that CenterPoint approach generally performs better than SECOND or PointPillars for the proposed dataset, this could be due to the nature of the approach where it deals directly with point clouds to predict object centers instead of voxelizing the points (SECOND) or projecting the point to BEV (PointPillars). We also provide results on the super-classes for the same architectures over different distance ranges and show similar performances for all approaches.

For the object tracking results presented in Table \ref{tab:exp4}, we notice a correlation between the detection scores and tracking scores (AP and AMOTA/MOTA) for classes such as Pedestrian and Car. We highlight that the detection as well as tracking models perform adequately on the proposed dataset achieving an overall AMOTA score of 0.472 (higher better), while we also note that a similar configuration achieves an overall AMOTA of 0.668 on the nuscenes dataset (from the leaderboard). The complexity of the proposed dataset is especially highlighted in the results of the Pedestrian class, where the low scores prove complex motion present in the dataset. We provide further results in the supplementary section along with the tracking results using SECOND and PointPillars models for completeness, with the corresponding visualizations.
\section{Conclusion}
In this work, we presented {IDD-3D}, a dataset for unstructured driving scenarios with complex road situations is presented with thorough statistical and experimental analysis. Through this dataset, and the future release, we aim to solve the problem of generalizability across geographical locations and provide more diverse information in driving datasets and road scene analysis. We show interesting cases which cover a manifold of cases but also show some safety-critical situations which are frequent in several cities. We justify our claims for the proposed dataset through a set of experiments for 3D object detection and tracking using state-of-the-art approaches which were available as open-source implementations. The future works for the dataset shall extend these tasks to a vast number of applications, further enhancing the applicability of the proposed dataset to autonomous driving applications.

\section{Acknowledgements}
The project is funded by iHub-data and mobility at IIIT Hyderabad. The authors would like to acknowledge the support from Radha Krishna B towards data collection and annotation. We would also like to thank Government of Telangana for the permissions, encouragement and enabling this effort.

{\small
\bibliographystyle{ieee_fullname}
\bibliography{egbib}

\begin{thebibliography}{10}\itemsep=-1pt

\bibitem{bao2020uncertainty}
Wentao Bao, Qi Yu, and Yu Kong.
\newblock Uncertainty-based traffic accident anticipation with spatio-temporal
  relational learning.
\newblock In {\em Proceedings of the 28th ACM International Conference on
  Multimedia}, pages 2682--2690, 2020.

\bibitem{brostow2009semantic}
Gabriel~J Brostow, Julien Fauqueur, and Roberto Cipolla.
\newblock Semantic object classes in video: A high-definition ground truth
  database.
\newblock {\em Pattern Recognition Letters}, 30(2):88--97, 2009.

\bibitem{nuscenes}
Holger Caesar, Varun Bankiti, Alex~H. Lang, Sourabh Vora, Venice~Erin Liong,
  Qiang Xu, Anush Krishnan, Yu Pan, Giancarlo Baldan, and Oscar Beijbom.
\newblock nuscenes: A multimodal dataset for autonomous driving.
\newblock In {\em CVPR}, 2020.

\bibitem{chandra2021meteor}
Rohan Chandra, Mridul Mahajan, Rahul Kala, Rishitha Palugulla, Chandrababu
  Naidu, Alok Jain, and Dinesh Manocha.
\newblock Meteor: A massive dense \& heterogeneous behavior dataset for
  autonomous driving.
\newblock {\em arXiv preprint arXiv:2109.07648}, 2021.

\bibitem{chang2019argoverse}
Ming-Fang Chang, John Lambert, Patsorn Sangkloy, Jagjeet Singh, Slawomir Bak,
  Andrew Hartnett, De Wang, Peter Carr, Simon Lucey, Deva Ramanan, et~al.
\newblock Argoverse: 3d tracking and forecasting with rich maps.
\newblock In {\em Proceedings of the IEEE/CVF Conference on Computer Vision and
  Pattern Recognition}, pages 8748--8757, 2019.

\bibitem{chen2017multi}
Xiaozhi Chen, Huimin Ma, Ji Wan, Bo Li, and Tian Xia.
\newblock Multi-view 3d object detection network for autonomous driving.
\newblock In {\em Proceedings of the IEEE conference on Computer Vision and
  Pattern Recognition}, pages 1907--1915, 2017.

\bibitem{choi2018kaist}
Yukyung Choi, Namil Kim, Soonmin Hwang, Kibaek Park, Jae~Shin Yoon, Kyounghwan
  An, and In~So Kweon.
\newblock Kaist multi-spectral day/night data set for autonomous and assisted
  driving.
\newblock {\em IEEE Transactions on Intelligent Transportation Systems},
  19(3):934--948, 2018.

\bibitem{collins2008towards}
Brendan Collins, Jia Deng, Kai Li, and Li Fei-Fei.
\newblock Towards scalable dataset construction: An active learning approach.
\newblock In {\em European conference on computer vision}, pages 86--98.
  Springer, 2008.

\bibitem{Cordts2016Cityscapes}
Marius Cordts, Mohamed Omran, Sebastian Ramos, Timo Rehfeld, Markus Enzweiler,
  Rodrigo Benenson, Uwe Franke, Stefan Roth, and Bernt Schiele.
\newblock The cityscapes dataset for semantic urban scene understanding.
\newblock In {\em Proc. of the IEEE Conference on Computer Vision and Pattern
  Recognition (CVPR)}, 2016.

\bibitem{Cordts2015Cvprw}
Marius Cordts, Mohamed Omran, Sebastian Ramos, Timo Scharw{\"a}chter, Markus
  Enzweiler, Rodrigo Benenson, Uwe Franke, Stefan Roth, and Bernt Schiele.
\newblock The cityscapes dataset.
\newblock In {\em CVPR Workshop on The Future of Datasets in Vision}, 2015.

\bibitem{dai2018dark}
Dengxin Dai and Luc Van~Gool.
\newblock Dark model adaptation: Semantic image segmentation from daytime to
  nighttime.
\newblock In {\em 2018 21st International Conference on Intelligent
  Transportation Systems (ITSC)}, pages 3819--3824. IEEE, 2018.

\bibitem{deng2020retinaface}
Jiankang Deng, Jia Guo, Evangelos Ververas, Irene Kotsia, and Stefanos
  Zafeiriou.
\newblock Retinaface: Single-shot multi-level face localisation in the wild.
\newblock In {\em Proceedings of the IEEE/CVF conference on computer vision and
  pattern recognition}, pages 5203--5212, 2020.

\bibitem{gahlert2020cityscapes}
Nils G{\"a}hlert, Nicolas Jourdan, Marius Cordts, Uwe Franke, and Joachim
  Denzler.
\newblock Cityscapes 3d: Dataset and benchmark for 9 dof vehicle detection.
\newblock {\em arXiv preprint arXiv:2006.07864}, 2020.

\bibitem{geiger2013vision}
Andreas Geiger, Philip Lenz, Christoph Stiller, and Raquel Urtasun.
\newblock Vision meets robotics: The kitti dataset.
\newblock {\em The International Journal of Robotics Research},
  32(11):1231--1237, 2013.

\bibitem{geiger2012we}
Andreas Geiger, Philip Lenz, and Raquel Urtasun.
\newblock Are we ready for autonomous driving? the kitti vision benchmark
  suite.
\newblock In {\em 2012 IEEE conference on computer vision and pattern
  recognition}, pages 3354--3361. IEEE, 2012.

\bibitem{grigorescu2020survey}
Sorin Grigorescu, Bogdan Trasnea, Tiberiu Cocias, and Gigel Macesanu.
\newblock A survey of deep learning techniques for autonomous driving.
\newblock {\em Journal of Field Robotics}, 37(3):362--386, 2020.

\bibitem{8575295}
X. {Huang}, X. {Cheng}, Q. {Geng}, B. {Cao}, D. {Zhou}, P. {Wang}, Y. {Lin},
  and R. {Yang}.
\newblock The apolloscape dataset for autonomous driving.
\newblock In {\em 2018 IEEE/CVF Conference on Computer Vision and Pattern
  Recognition Workshops (CVPRW)}, pages 1067--10676, 2018.

\bibitem{iyer2018calibnet}
Ganesh Iyer, R~Karnik Ram, J~Krishna Murthy, and K~Madhava Krishna.
\newblock Calibnet: Geometrically supervised extrinsic calibration using 3d
  spatial transformer networks.
\newblock In {\em 2018 IEEE/RSJ International Conference on Intelligent Robots
  and Systems (IROS)}, pages 1110--1117. IEEE, 2018.

\bibitem{jiang2021rellis}
Peng Jiang, Philip Osteen, Maggie Wigness, and Srikanth Saripalli.
\newblock Rellis-3d dataset: Data, benchmarks and analysis.
\newblock In {\em 2021 IEEE international conference on robotics and automation
  (ICRA)}, pages 1110--1116. IEEE, 2021.

\bibitem{kim2021eagermot}
Aleksandr Kim, Aljo{\v{s}}a O{\v{s}}ep, and Laura Leal-Taix{\'e}.
\newblock Eagermot: 3d multi-object tracking via sensor fusion.
\newblock In {\em 2021 IEEE International Conference on Robotics and Automation
  (ICRA)}, pages 11315--11321. IEEE, 2021.

\bibitem{lang2019pointpillars}
Alex~H Lang, Sourabh Vora, Holger Caesar, Lubing Zhou, Jiong Yang, and Oscar
  Beijbom.
\newblock Pointpillars: Fast encoders for object detection from point clouds.
\newblock In {\em Proceedings of the IEEE/CVF conference on computer vision and
  pattern recognition}, pages 12697--12705, 2019.

\bibitem{laroca2021efficient}
Rayson Laroca, Luiz~A Zanlorensi, Gabriel~R Gon{\c{c}}alves, Eduardo Todt,
  William~Robson Schwartz, and David Menotti.
\newblock An efficient and layout-independent automatic license plate
  recognition system based on the yolo detector.
\newblock {\em IET Intelligent Transport Systems}, 15(4):483--503, 2021.

\bibitem{li2020sustech}
E Li, Shuaijun Wang, Chengyang Li, Dachuan Li, Xiangbin Wu, and Qi Hao.
\newblock Sustech points: A portable 3d point cloud interactive annotation
  platform system.
\newblock In {\em 2020 IEEE Intelligent Vehicles Symposium (IV)}, pages
  1108--1115. IEEE, 2020.

\bibitem{liao2021kitti}
Yiyi Liao, Jun Xie, and Andreas Geiger.
\newblock Kitti-360: A novel dataset and benchmarks for urban scene
  understanding in 2d and 3d.
\newblock {\em arXiv preprint arXiv:2109.13410}, 2021.

\bibitem{liu2022bevfusion}
Zhijian Liu, Haotian Tang, Alexander Amini, Xinyu Yang, Huizi Mao, Daniela Rus,
  and Song Han.
\newblock Bevfusion: Multi-task multi-sensor fusion with unified bird's-eye
  view representation.
\newblock {\em arXiv preprint arXiv:2205.13542}, 2022.

\bibitem{mao2021one}
Jiageng Mao, Minzhe Niu, Chenhan Jiang, Hanxue Liang, Jingheng Chen, Xiaodan
  Liang, Yamin Li, Chaoqiang Ye, Wei Zhang, Zhenguo Li, et~al.
\newblock One million scenes for autonomous driving: Once dataset.
\newblock {\em arXiv preprint arXiv:2106.11037}, 2021.

\bibitem{pandey2012automatic}
Gaurav Pandey, James~R McBride, Silvio Savarese, and Ryan~M Eustice.
\newblock Automatic targetless extrinsic calibration of a 3d lidar and camera
  by maximizing mutual information.
\newblock In {\em Twenty-Sixth AAAI Conference on Artificial Intelligence},
  2012.

\bibitem{pang2021simpletrack}
Ziqi Pang, Zhichao Li, and Naiyan Wang.
\newblock Simpletrack: Understanding and rethinking 3d multi-object tracking.
\newblock {\em arXiv preprint arXiv:2111.09621}, 2021.

\bibitem{pham20203d}
Quang-Hieu Pham, Pierre Sevestre, Ramanpreet~Singh Pahwa, Huijing Zhan, Chun~Ho
  Pang, Yuda Chen, Armin Mustafa, Vijay Chandrasekhar, and Jie Lin.
\newblock A\textasteriskcentered 3d dataset: Towards autonomous driving in
  challenging environments.
\newblock In {\em 2020 IEEE International Conference on Robotics and Automation
  (ICRA)}, pages 2267--2273. IEEE, 2020.

\bibitem{pitropov2021canadian}
Matthew Pitropov, Danson~Evan Garcia, Jason Rebello, Michael Smart, Carlos
  Wang, Krzysztof Czarnecki, and Steven Waslander.
\newblock Canadian adverse driving conditions dataset.
\newblock {\em The International Journal of Robotics Research},
  40(4-5):681--690, 2021.

\bibitem{qi2018frustum}
Charles~R Qi, Wei Liu, Chenxia Wu, Hao Su, and Leonidas~J Guibas.
\newblock Frustum pointnets for 3d object detection from rgb-d data.
\newblock In {\em Proceedings of the IEEE conference on computer vision and
  pattern recognition}, pages 918--927, 2018.

\bibitem{quigley2009ros}
Morgan Quigley, Ken Conley, Brian Gerkey, Josh Faust, Tully Foote, Jeremy
  Leibs, Rob Wheeler, Andrew~Y Ng, et~al.
\newblock Ros: an open-source robot operating system.
\newblock In {\em ICRA workshop on open source software}, volume~3, page~5.
  Kobe, Japan, 2009.

\bibitem{rebecq2019high}
Henri Rebecq, Ren{\'e} Ranftl, Vladlen Koltun, and Davide Scaramuzza.
\newblock High speed and high dynamic range video with an event camera.
\newblock {\em IEEE transactions on pattern analysis and machine intelligence},
  43(6):1964--1980, 2019.

\bibitem{ren2021survey}
Pengzhen Ren, Yun Xiao, Xiaojun Chang, Po-Yao Huang, Zhihui Li, Brij~B Gupta,
  Xiaojiang Chen, and Xin Wang.
\newblock A survey of deep active learning.
\newblock {\em ACM Computing Surveys (CSUR)}, 54(9):1--40, 2021.

\bibitem{sakaridis2021acdc}
Christos Sakaridis, Dengxin Dai, and Luc Van~Gool.
\newblock Acdc: The adverse conditions dataset with correspondences for
  semantic driving scene understanding.
\newblock In {\em Proceedings of the IEEE/CVF International Conference on
  Computer Vision}, pages 10765--10775, 2021.

\bibitem{sharma2022cat}
Suvash Sharma, Lalitha Dabbiru, Tyler Hannis, George Mason, Daniel~W Carruth,
  Matthew Doude, Chris Goodin, Christopher Hudson, Sam Ozier, John~E Ball,
  et~al.
\newblock Cat: Cavs traversability dataset for off-road autonomous driving.
\newblock {\em IEEE Access}, 10:24759--24768, 2022.

\bibitem{sindagi2019mvx}
Vishwanath~A Sindagi, Yin Zhou, and Oncel Tuzel.
\newblock Mvx-net: Multimodal voxelnet for 3d object detection.
\newblock In {\em 2019 International Conference on Robotics and Automation
  (ICRA)}, pages 7276--7282. IEEE, 2019.

\bibitem{sun2020scalability}
Pei Sun, Henrik Kretzschmar, Xerxes Dotiwalla, Aurelien Chouard, Vijaysai
  Patnaik, Paul Tsui, James Guo, Yin Zhou, Yuning Chai, Benjamin Caine, et~al.
\newblock Scalability in perception for autonomous driving: Waymo open dataset.
\newblock In {\em Proceedings of the IEEE/CVF conference on computer vision and
  pattern recognition}, pages 2446--2454, 2020.

\bibitem{8659045}
Girish Varma, Anbumani Subramanian, Anoop Namboodiri, Manmohan Chandraker, and
  C.V. Jawahar.
\newblock Idd: A dataset for exploring problems of autonomous navigation in
  unconstrained environments.
\newblock In {\em 2019 IEEE Winter Conference on Applications of Computer
  Vision (WACV)}, pages 1743--1751, 2019.

\bibitem{weng20203d}
Xinshuo Weng, Jianren Wang, David Held, and Kris Kitani.
\newblock 3d multi-object tracking: A baseline and new evaluation metrics.
\newblock In {\em 2020 IEEE/RSJ International Conference on Intelligent Robots
  and Systems (IROS)}, pages 10359--10366. IEEE, 2020.

\bibitem{wigness2019rugd}
Maggie Wigness, Sungmin Eum, John~G Rogers, David Han, and Heesung Kwon.
\newblock A rugd dataset for autonomous navigation and visual perception in
  unstructured outdoor environments.
\newblock In {\em 2019 IEEE/RSJ International Conference on Intelligent Robots
  and Systems (IROS)}, pages 5000--5007. IEEE, 2019.

\bibitem{wilson2021argoverse}
Benjamin Wilson, William Qi, Tanmay Agarwal, John Lambert, Jagjeet Singh,
  Siddhesh Khandelwal, Bowen Pan, Ratnesh Kumar, Andrew Hartnett,
  Jhony~Kaesemodel Pontes, et~al.
\newblock Argoverse 2: Next generation datasets for self-driving perception and
  forecasting.
\newblock 2021.

\bibitem{yan2018second}
Yan Yan, Yuxing Mao, and Bo Li.
\newblock Second: Sparsely embedded convolutional detection.
\newblock {\em Sensors}, 18(10):3337, 2018.

\bibitem{yang2018pixor}
Bin Yang, Wenjie Luo, and Raquel Urtasun.
\newblock Pixor: Real-time 3d object detection from point clouds.
\newblock In {\em Proceedings of the IEEE conference on Computer Vision and
  Pattern Recognition}, pages 7652--7660, 2018.

\bibitem{yin2021center}
Tianwei Yin, Xingyi Zhou, and Philipp Krahenbuhl.
\newblock Center-based 3d object detection and tracking.
\newblock In {\em Proceedings of the IEEE/CVF conference on computer vision and
  pattern recognition}, pages 11784--11793, 2021.

\bibitem{yogamani2019woodscape}
Senthil Yogamani, Ciar{\'a}n Hughes, Jonathan Horgan, Ganesh Sistu, Padraig
  Varley, Derek O'Dea, Michal Uric{\'a}r, Stefan Milz, Martin Simon, Karl
  Amende, et~al.
\newblock Woodscape: A multi-task, multi-camera fisheye dataset for autonomous
  driving.
\newblock In {\em Proceedings of the IEEE/CVF International Conference on
  Computer Vision}, pages 9308--9318, 2019.

\bibitem{yoo2019learning}
Donggeun Yoo and In~So Kweon.
\newblock Learning loss for active learning.
\newblock In {\em Proceedings of the IEEE/CVF Conference on Computer Vision and
  Pattern Recognition}, pages 93--102, 2019.

\bibitem{9046805}
E. {Yurtsever}, J. {Lambert}, A. {Carballo}, and K. {Takeda}.
\newblock A survey of autonomous driving: Common practices and emerging
  technologies.
\newblock {\em IEEE Access}, 8:58443--58469, 2020.

\bibitem{zhou2018voxelnet}
Yin Zhou and Oncel Tuzel.
\newblock Voxelnet: End-to-end learning for point cloud based 3d object
  detection.
\newblock In {\em Proceedings of the IEEE conference on computer vision and
  pattern recognition}, pages 4490--4499, 2018.

\end{thebibliography}
}

\end{document}